\documentclass[conference, compsoc]{IEEEtran}
\usepackage[backend=bibtex,bibstyle=ieee,citestyle=numeric-comp]{biblatex}
\AtBeginBibliography{\small}
\bibliography{bibliography}
\IEEEoverridecommandlockouts
\usepackage{amsmath,amssymb,amsfonts}
\usepackage{algorithmic}
\usepackage{graphicx}
\usepackage{textcomp}
\usepackage{xcolor}
\usepackage{mathtools}
\usepackage{dsfont}
\usepackage{tabularx}
\newcolumntype{Y}{>{\centering\arraybackslash}X}
\usepackage{caption} 
\usepackage{subcaption}
\usepackage{mathrsfs}  
\usepackage{todonotes}
\usepackage[normalem]{ulem}

\newcolumntype{P}[1]{>{\centering\arraybackslash}p{#1}}

\begin{document}

\title{Who breaks early, looses: goal oriented training of  deep neural networks  based on port Hamiltonian dynamics}

\author{\IEEEauthorblockN{Julian Burghoff}
\IEEEauthorblockA{\textit{Department of Mathematics \& IZMD} \\
\textit{University of Wuppertal}\\
Wuppertal, Germany \\
burghoff@math.uni-wuppertal.de}

\and

\IEEEauthorblockN{Marc Heinrich Monells}
\IEEEauthorblockA{\textit{Department of Mathematics} \\
\textit{University of Wuppertal}\\
Wuppertal, Germany \\
marc.heinrich\_monells@uni-wuppertal.de}

\and

\IEEEauthorblockN{Hanno Gottschalk}
\IEEEauthorblockA{\textit{Institute of Mathematics} \\
\textit{TU-Berlin}\\
Berlin, Germany \\
gottschalk@math.tu-berlin.de}
}

\maketitle

\begin{abstract}
The highly structured energy landscape of the loss as a function of parameters for deep neural networks makes it necessary to use sophisticated optimization strategies in order to discover  (local) minima that guarantee reasonable performance. Overcoming less suitable local minima is an important prerequisite and often momentum methods are employed to achieve this. As in other non local optimization procedures, this however creates the necessity to balance between exploration and exploitation.  In this work, we suggest an event based control mechanism for switching from exploration to exploitation based on reaching a predefined reduction of the loss function. As we give the momentum method a  port Hamiltonian interpretation, we apply the 'heavy ball with friction' interpretation and trigger breaking (or friction) when achieving certain goals. We benchmark our method against standard stochastic gradient descent and provide experimental evidence for improved performance of deep neural networks when our strategy is applied.  
\end{abstract}


\section{Introduction}
The success of deep neural networks (DNN) significantly depends on the cheap computation of gradients using back-propagation enabling gradient based minimization of the loss functions. As the parameter count of DNN ranges between several tens of thousand in small classification networks to several billion in large scale generative models, there seems to be no alternative to the use of gradients. However, gradient based optimization is beset with the problem of local minima, of which the energy landscape of DNN offers plenty. Exploitation of a local minimum with gradient descent comes with guarantees for progress relative to previous optimization steps, but does not guarantee a decent level of performance. In order to go more global, momentum methods have therefore been introduced to overcome local minima.    

As compared to gradient descent, momentum based methods have more parameters to adjust. Besides the strength of the inertial forces controlled by the 'mass' parameter, a 'friction' parameter has to be determined, which is responsible for slowing down the search motion and bringing it to rest, ultimately. Finally, the learning rate needs to be controlled throughout the progress of the optimization process, like in gradient descent. 

The complexity in setting and controlling the aforementioned hyper-parameters can be alleviated by an interpretation of the optimization process in physical terms as already indicated by the physical connotations of 'mass' and 'friction'. It has been recently proposed to cast the optimization process in a port Hamiltonian framework, which makes the  convergence of the optimization process to a stationary point transparent via energy based considerations, where loss is connected to potential and momentum to kinetic energy, whereas 'friction' accounts for energy dissipation and interdicts motion at high pace for unlimited time.     
It is clear that the friction / energy dissipation parameter is essential for the (non) locality of the optimization process: if high, friction essentially damps out all momentum and the procedure essentially 'just flows down the hill' as for gradient descent, resulting in low exploration and high exploitation. If low, the motion will go on essentially un-damped and not rest and thereby explore all of the accessible parameter space. Exploration is high, and exploitation is low in this setting.

Then, parameter settings can be modified over time or controlled adaptively as a part of the optimization algorithm is a familiar thought.  The physics based intuition of port Hamiltonian systems can be helpful in the design of such adaptive strategies. Here we suggest a simple, event based adaptive parameter selection strategy that starts the optimization in an exploratory phase with low friction and turns over to exploitation by 'heavy breaking', once the potential energy (i.e.\ the loss function) is sufficiently reduced.  Sufficiency is pre-defined as the minimum reduction goal of the optimization, which can be set, e.g., as the reduction of the loss obtained in previous trials. 

In this paper, we show that the proposed strategy actually works for some classical examples in deep learning and improves the optimization loss and also the test accuracy for a standard, Le-Net-5 \cite{lecun1998gradient} based architectures on two well known academic classification tasks solved by deep learning, namely the CIFAR10 \cite{CIFAR10} and the FashionMNIST \cite{FashionMNIST} data-sets. 

In order to focus on the optimization only, we do not employ data augmentation or pre-training and thereby do not achieve SOTA performance in our experiments. We however consistently achieve an advantage over the widely used stochastic gradient descent as a benchmark. We also observe consistent gains in performance after 'heavy breaking' is finally triggered.

Our paper is organized as follows: in Section \ref{sec:RelatedWork} we give an overview over related work and in Section \ref{sec:methods} we present the port Hamiltonian view on gradient based optimization with momentum and energy dissipation. Our experimental setup as well as our results are documented in Section \ref{sec:experiments}.
In the final Section \ref{sec:discussion} we present our conclusions and give an outlook to future research.


\section{Related Work}
\label{sec:RelatedWork}

The fact that neural networks with parameter counts ranging from some tenth of thousands  to several hundreds of billions can actually be trained, largely depends on the cheap computation of gradients, see \cite{werbos2005applications,lecun1998MNIST} for original work and \cite{goodfellow2016deep} for a recent reference. Gradient based optimization itself has been studied since the days of Newton, see e.g.\ \cite{baza:nonl:2006,wright1999numerical}.  In the context of deep learning, the formation of randomly sub-sampled mini-batches  is necessary as big data often exceeds the working memory available \cite{li2014efficient}. One has therefore to pass over to the stochastic gradient descent method (SGD) \cite{saad1998online,shalev2014understanding}.

One of the problems in neural network training is the complex, non convex structure of the energy landscapes \cite{becker2020geometry}. This makes it necessary to avoid local minima, which is mostly done by the momentum method \cite{nesterov1983method,goh2017momentum,qian1999momentum}. From a theoretical side, momentum can be understood as a discretized version of a second order ordinary differential equation, which also provides theoretical insight to convergence to critical points \cite{anti:seco:1993,atto:fast:2018,poly:some:1964}, see also \cite{ochs:ipia:2014,ochs:loca:2018,ochs:adap:2019} for recent extensions. 

The momentum method has recently be cast in a modern port Hamiltonian language \cite{massaroli2019port,poli2020port,kova:cont:2021}. Port Hamiltonian systems \cite{van2014port} are particularly suited to understand the long time behviour and hence convergence properties of momentum based methods.

For a long time, the control of hyperparemeters in the training of neural networks has been a topic of interest in the deep learning community  \cite{bengio2012practical}. While learning rate schedules \cite{darken1990note,darken1992learning} determine the setting for one specific parameter upfront, it has also been proposed to modify the dissipation parameter in momentum based optimization \cite{cabo:onth:2009,atto:fast:2018,cham:onth:2015}. Other strategies, like the much used ADAM algorithm, rely on adaptive parameter control \cite{kingma2014adam,bock2019proof}. 

One specific adaptive strategy however much less considered is the goal oriented search, where one pre-defines the target value to achieve during optimization, see e.g. \cite{sobester2008engineering}.  

In our work, we thus make the following contributions: 
\begin{itemize}
    \item For the first time, we use the port Hamiltonian language in the training of reasonably \emph{deep} neural networks in contracst to \cite{massaroli2019port,poli2020port} where networks are shallow.
    \item We also introduce an adaptive, goal oriented strategy for the control of the friction constant, which goes in the opposite direction as  \cite{cabo:onth:2009,atto:fast:2018,cham:onth:2015} but is well-motivated in terms of combining exploration and exploitation in one algorithm.
    \item We show experimentally for standard deep learning problems in image recognition that this strategy consistently produces improvements over fixed-parameter strategies. 
    We also provide a considerable amount of ablation studies related to our parameter settings. 
\end{itemize}

\section{The Goal Oriented PHS Method}
\label{sec:methods}
The simple gradient descent algorithm to minimize a differentiable loss function $\mathscr{L}(\theta)$, namely $\theta_{k+1}=\theta_k-\alpha \nabla_\theta \mathscr{L}(\theta_k)$ can be seen as a first order Euler discretization of the gradient flow 
\begin{equation}
    \label{eq:gradientFlow}
    \dot \theta(t)=-\nabla_\theta \mathscr{L}(\theta),~~\theta(0)=\theta_0.
\end{equation}
It is well known that under adequate conditions on $\mathscr{L}(\theta)$, the flow $\theta(t)$ converges for $t\to\infty$ to a critical point $\theta^*$ with $\nabla_\theta \mathscr{L}(\theta^*)=0$, see e.g. \cite{massaroli2019port,poli2020port}. Likewise, the gradient descent algorithm converges for $k\to\infty$ to a critical point, provided the step length $\alpha$ is suitably controlled, confer \cite{anti:seco:1993,atto:fast:2018}.

As mentioned in the introduction, the problem with gradient descent in the context of highly non-convex loss functions $\mathscr{L}(\theta)$, as especially in the context of the training of deep neural networks \cite{goodfellow2016deep}, lies in the fact that gradient flows and gradient descent algorithms get stuck in local minima.  

To over come the strict locality of gradient flow and gradient descent, momentum based methods have been introduced. The update rule of gradient descent is changed to
\begin{align}
    \label{eq:momentum}
    \begin{split}
    \theta_{k+1}&=\theta_k+\alpha \frac{1}{m} p_k\\
    p_{k+1}&=p_k-\alpha \frac{\gamma}{m} p_k -\alpha \nabla_\theta \mathscr{L}(\theta)
    \end{split}
\end{align}
where $m,\gamma>0$ are parameters called mass and friction coefficient.  $p_k$ is the so-called momentum at iteration $k$. In fact, \eqref{eq:momentum} can be understood as the discretized version of the following Hamiltonian set of equations
\begin{align}
    \label{eq:Hamiltonian_Dynamics}
    \begin{split}
    \dot \theta(t)&= \frac{1}{m} p(t)\\
    \dot p(t)&=- \frac{\gamma}{m} p(t) - \nabla_\theta \mathscr{L}(\theta)
    \end{split}
\end{align}
    with initial conditions $\theta(0)=\theta_0$ and $p(0)=p_0$. 

    To understand the global properties of the Hamiltonian dynamics, it is convenient to define a state variable $x(t)=\left({\theta(t)\atop p(t)}\right)$ and the Hamiltonial function $H(x)=\frac{\|p\|^2}{2m}+\mathscr{L}(\theta)$ and a the symplectic matrix $J=\left(\begin{array}{cc}
         0&-1  \\
         1&0 
    \end{array}\right)$ as well as a symmetric, positive resistive matrix $J=\left(\begin{array}{cc}
         0&0  \\
         0&\frac{\gamma}{m} 
    \end{array}\right)$ so that we can rewrite \eqref{eq:Hamiltonian_Dynamics} in the compact, port-Hamiltonian form
    \begin{equation}
        \label{eq:pHS}
        \dot x(t)=\left(J-R\right)\nabla_{x}H(x).
    \end{equation}
    Using the chain-rule, \eqref{eq:pHS} and $\nabla_{x}H(x(\tau))^\top J \nabla_{x}H(x(\tau))=0$ by the skew-symmetry of $J$, it is now easy to see that the following inequality holds for the dissipated total 'energy' measured by $H(x)$, where $\frac{\|p\|^2}{2m}$ takes the role of kinetic energy and the loss $\mathscr{L}(\theta)$ the role of potential energy   
\begin{equation}
\label{eq:dissipation}
    H(x(t))-H(x(0))=-\int_0^t \nabla_{x}H(x(\tau))^\top R \nabla_{x}H(x(\tau))\, \mathrm{d} \tau. 
\end{equation}
 From this exposition it is intuitive, and in fact can be proven mathematically \cite{anti:seco:1993,atto:fast:2018}, that due to dissipation the state $x(t)$ ultimately has to come to a rest, if $\mathscr{L}(\theta)$ is bounded from below.  Thus, if the stationary points $x^*$ with $\nabla_xH(x^*)=0$ of the system are isolated, $x(t)$ will asymptotically converge to a stationary point. Furthermore, for $x^*=\left({\theta^*\atop p^*}\right)$, we find $p^*=0$ and $\nabla_\theta \mathscr{L}(\theta^*)=0$, hence the $\theta$-component of stationary points are in one to one correspondence to the critical points of the original optimization problem. 
 
Energy dissipation \eqref{eq:dissipation} thus is the key component that determines how fast $x(t)$ comes to rest, which conceptually is corresponding to convergence of the optimization algorithm. Apparently, the matrix $R$ and thus the friction coefficient $\gamma$ controls dissipation.

In fact, if $\gamma\approx 0$, essentially no energy is lost and the dynamics $x(t)$ will either move on for a very long time, or, in very rare cases, get to rest on a local maximum or saddle point. This perpetual motion through the accessible part of the 'phase space' can be seen as an exploitative strategy.

In contrast, if $\gamma$ gets large, the friction essentially disperses energy and momentum and the motion of $x(t)$ behaves highly viscous, i.e. determined by the equality 
\begin{equation}
    \label{eq:supercritical}
    -\frac{\gamma}{m} p(t)-\nabla_\theta \mathscr{L}(\theta)\approx 0~~\Leftrightarrow ~~ \dot \theta(t) \approx -\frac{1}{\gamma} \nabla_\theta \mathscr{L}(\theta),
\end{equation}
from which we see that in this high viscosity regime the port Hamiltonian flow essentially behaves like gradient descent (with a modified step length). Despite working with momentum, we are thus back in the exploitation phase of local minima.

The idea of this article is to use this physics based intuition to efficiently control the behavior of our port Hamiltonian optimization strategy in a goal oriented search. We thus propose to 'keep on moving' as long as we have not yet reached a predefined reduction of the initial loss function $\mathscr{L}(\theta_0)$. In many cases,  it is known that $\mathscr{L}(\theta)$ is lower bounded by zero, and we can thus demand a $90\%$, $95\%$ \ldots reduction in $\mathscr{L}(x(t))$, before we, upon reaching this target, instantaneously increase the value of $\gamma$ in order to switch over from the low-viscous exploration phase to high-viscous exploitation. In this sense, our proposed optimization algorithm resembles the 'chicken game': who breaks too early, looses.

Before we come to the implementation and numerical tests of this strategy in deep learning, we discuss some peculiarities of the loss function in this case. We would like to learn a conditional probability density $p(y|x,\theta)$ from data independently sampled from the same distribution $\{(y_i,x_i)\}_{i=1}^n$, where $x_i$ is some input and $y_i$ takes values in some prescribed label space $\mathscr{C}=\{c_1,\ldots,c_q\}$. In applications in image recognition, $p(y|x,\theta)$ often consists of several stacked convolutional and fully connected layers and an ultimate softmax layer, cf.\ \cite{goodfellow2016deep}. The 'cross entropy'/negative log likelihood  loss is given by
\begin{equation}
    \label{eq:ceLoss}
    \mathscr{L}(\theta)=-\frac{1}{n}\sum_{i=1}^n\log p(y_i|x_i,\theta).
\end{equation}
The numerical problem to implement \eqref{eq:ceLoss} directly lies in the memory constraints that do not permit to load the entire data set $\{(y_i,x_i)\}_{i=1}^n$ in the working memory. Therefore, mini batches $B_j$, i.e.\ small random subsets of $\{1,\ldots,n\}$ are drawn and an update step of the parameters $\theta_k$ and the associated momentum is executed for a loss $\mathscr{L}_{B_j}(\theta)$ with the original data set replaced by $\{(y_i,x_i)\}_{i\in B_j}$. Nevertheless, as in image classification oftentimes the batch $|B_j|$ is quite large ($\gtrapprox 10$), $\mathscr{L}_{B_j}(\theta)$ and $\mathscr{L}(\theta)$ tend do behave similar by the law of large numbers. In our numerical experiments, we therefore observe the behavior of the algorithm in accordance with intuition. 

\section{Experiments and results}
\label{sec:experiments}

\begin{figure}
    \centering
    \includegraphics[width=\linewidth]{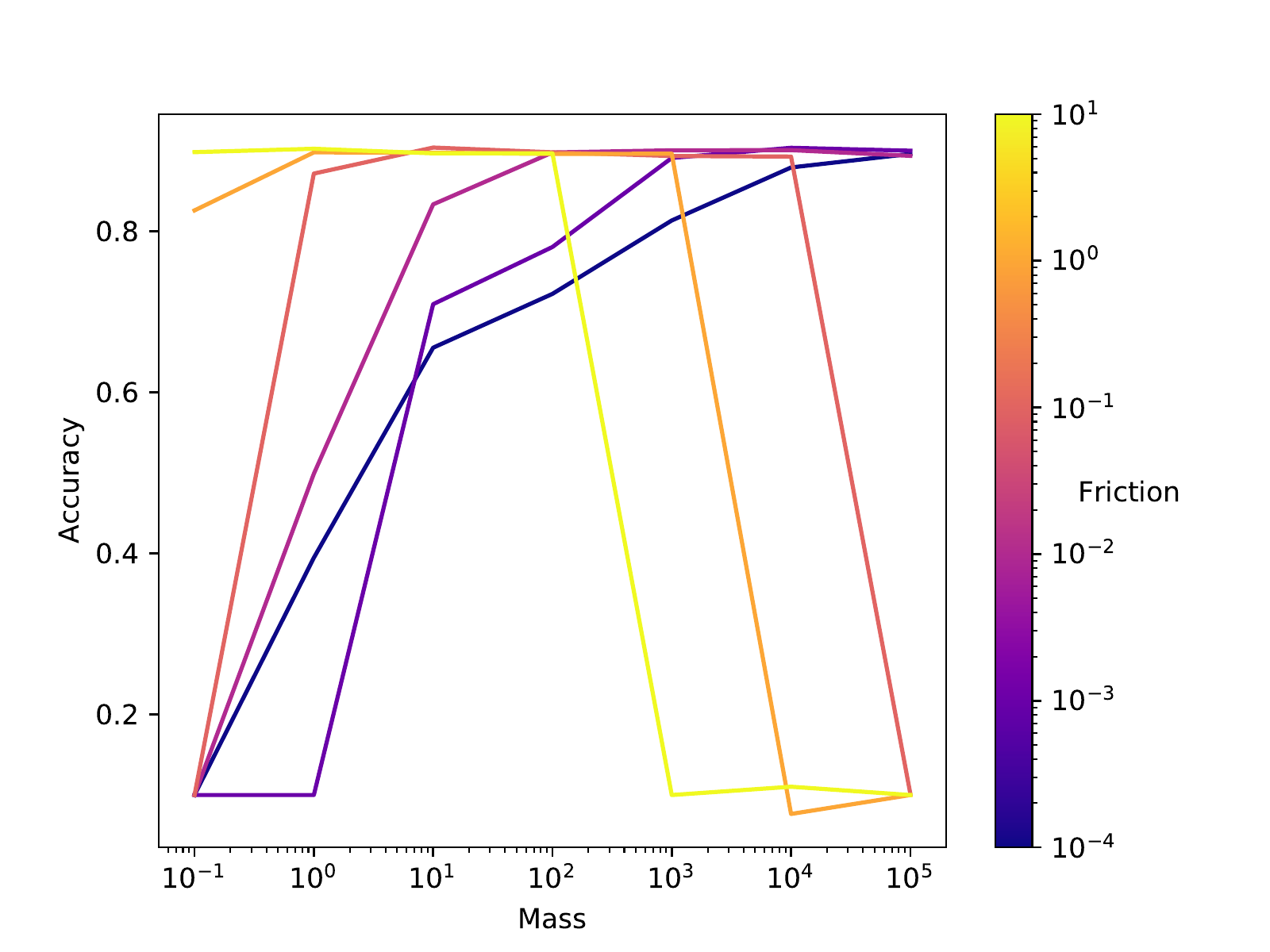}
    \caption{Selecting hyperparameters of learning rate (here: $\alpha = 0.1$), mass and friction based on the accuracy on the Fahion-MNIST dataset}
    \label{fig:fric_mass}
\end{figure}

\begin{figure}
    \centering
    \includegraphics[width=\linewidth]{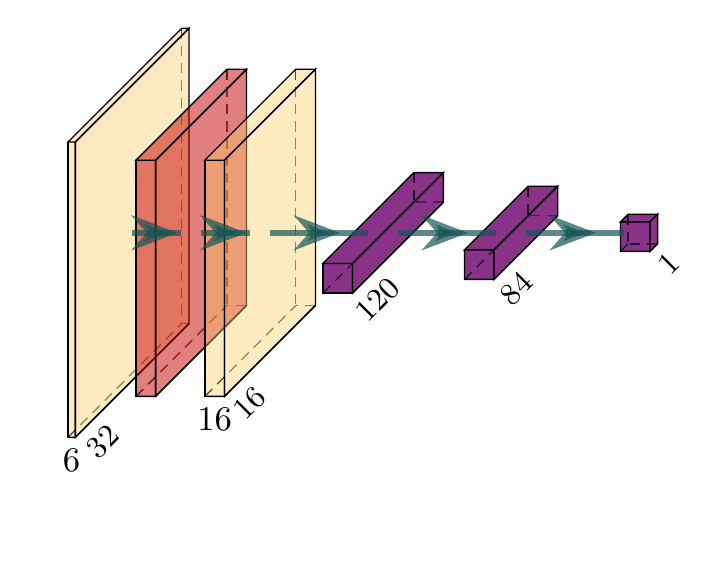}
    \caption{Neural Net architecture which is similar to Le-Net-5. Orange are convolutional layers with a filter size of 5, red is the pooling layer and fully connected layers are violet.}
    \label{fig:used_architecture}
\end{figure}

For our experiments, we use a Convolutional Neural Net (CNN) similar to the Le-Net-5\cite{lecun1998gradient} which consists of two convolutional, one pooling and two fully connected layers as it is shown in figure \ref{fig:used_architecture} and has a total of 44,426 weights. For implementation we are using the PyTorch framework \cite{pytorch2019}. 
This network is chosen as it is a widely used standard architecture, although it is not eligible to compete with more sophisticated ResNet \cite{he2016deep} or Transformer \cite{vaswani2017attention} architectures. Furthermore, in order to focus on training exclusively,  the networks are trained from scratch on the data sets and we use neither pre-training nor augmentation. The training is performed with respect to the usual cross-entropy loss without regularization. 

On the hardware-side, we use a workstation with an Intel(R) Core(TM) i7-6850K 3.6GHz and two Nvidia TITAN Xp graphic units with 12GB VRAM each for our experiments.

For a comparison with SGD and PHS, i.e. the traditional momentum method, we test our goal oriented PHS search on the two data sets CIFAR10 and FashionMNIST introduced above. We furthermore run trainings for a number of different learning rates $\alpha$ and for several settings for the mass and baseline friction parameter. To establish which parameter settings are rewarding, we consider the accuracies of the PHS for different learning rates ($0.0001 \leq \alpha \leq 0.1$), that can be achieved when mass and friction are included. This is shown in Figure \ref{fig:fric_mass} for the example of $\alpha=0.1$ on the Fashion-MNIST dataset. As one can already see, the trainings for many parameter settings work significantly worse or not at all. Therefore, only experiments that lie in a parameter range leading to reasonable results are included in our result tables. Concerning goal orientation, we aim at an reduction of the initial loss of 65\% to 90\% and then increase the friction significantly by a factor between 5 and 99. The results are given in  
Tables \ref{table:accuracies_cifar10}  for CIFAR10 and \ref{table:accuracies_fashionmnist} for FashionMNIST.

\begin{table}[]
\begin{tabular}{|c|P{3.5cm}|c|c|c|c|}
\hline
$\alpha$ & Optimizer & Fric & Mass & Acc\\
\hline
0.1 & SGD & / & / & 64.82\% \\
\hline
0.1 & PHS & 0.1 & 100 & 66.45\% \\
\hline
0.1 & Goal-Oriented (breaking at 0.65 with factor 49) & 0.1 & 100 & \textbf{67.1\%}  \\
\hline
0.1 & PHS & 0.01 & 100 & 63.52\% \\
\hline
0.1 & Goal-Oriented (breaking at 0.9 with factor 99) & 0.01 & 100 & 65.52\% \\
\hline
0.01 & SGD & / & / & 63.53\% \\
\hline
0.01 & PHS & 0.1 & 25 & 66.01\% \\
\hline
0.01 & Goal-Oriented (breaking at 0.7 with factor 10) & 0.1 & 25 &\textbf{66.49\%} \\
\hline
0.01 & PHS & 0.01 & 25 & 62.98\% \\
\hline
0.01 & Goal-Oriented (breaking at 0.7 with factor 50) & 0.01  & 25 & 63.44\% \\
\hline
0.001 & SGD & / & / & 65.05 \% \\
\hline
0.001 & PHS & 1 & 0.25 & 66.0 \% \\
\hline
0.001 & Goal-Oriented (breaking at 0.7 with factor 20) & 1  & 0.25 & \textbf{66.37\%}  \\
\hline
0.001 & PHS & 0.1 & 0.25 & 62.93\% \\
\hline
0.001 & Goal-Oriented (breaking at 0.85 with factor 50) & 0.1 & 0.25 & 63.54\%\\
\hline
0.0001 & SGD & / & / & 64.43\% \\
\hline
0.0001 & PHS & 10 & 0.001 & 65.76\% \\
\hline
0.0001 & Goal-Oriented (breaking at 0.68 with factor 5) & 10  & 0.001 & \textbf{66.39\%} \\
\hline
0.0001 & PHS & 1 & 0.001 & 62.24\% \\
\hline
0.0001 & Goal-Oriented (breaking at 0.8 with factor 100) & 1 & 0.001 & 63.56\% \\
\hline
\end{tabular}
\caption{Comparison of training results with SGD, PHS and Goal-Oriented approaches for the CIFAR-10 dataset.} \label{table:accuracies_cifar10}
\end{table}

\begin{table}[]
\begin{tabular}{|c|P{3.5cm}|c|c|c|c|}
\hline
$\alpha$ & Optimizer & Fric & Mass & Acc\\
\hline
0.1 & SGD & / & /  & 90.04\% \\
\hline
0.1 & PHS & 0.1 & 10  & 90.36\% \\
\hline
0.1 & Goal-Oriented (breaking at 0.15 with factor 50) & 0.1 & 10  & \textbf{91.02\%}\\
\hline
0.1 & PHS & 0.01 & 10 & 83.31\%\\
\hline
0.1 & Goal-Oriented (breaking at 0.55 with factor 20) & 0.01  & 10 & 87.19\% \\
\hline
0.01 & SGD & / & / & 90.26\% \\
\hline
0.01 & PHS & 1 & 0.1  & 90.49\% \\
\hline
0.01 & Goal-Oriented (breaking at 0.2 with factor 10) & 1 & 0.1  & \textbf{90.98\%}\\
\hline
0.01 & PHS & 0.1 & 0.1  & 83.28\% \\
\hline
0.01 & Goal-Oriented (breaking at 0.5 with factor 5) & 0.1  & 0.1  & 86.47\% \\
\hline
0.001 & SGD & / & /  & 89.61\% \\
\hline
0.001 & PHS & 10 & 0.01  & 90.34\% \\
\hline
0.001 & Goal-Oriented (breaking at 0.15 with factor 5) & 10 & 0.01  & \textbf{90.8\%}\\
\hline
0.001 & PHS & 1 & 0.01  & 90.13\% \\
\hline
0.001 & Goal-Oriented (breaking at 0.17 with factor 50) & 1  & 0.01  & 90.77\% \\
\hline
0.0001 & SGD & / & /  & 88.86\% \\
\hline
0.0001 & PHS & 10 & 0.001 & 90.17\% \\
\hline
0.0001 & Goal-Oriented (breaking at 0.2 with factor 100) & 10 & 0.001 & \textbf{90.54\%} \\
\hline
0.0001 & PHS & 1 & 0.001 & 89.6\% \\
\hline
0.0001 & Goal-Oriented (breaking at 0.185 with factor 100) & 1 & 0.001 & 90.12\% \\
\hline
\end{tabular}
\caption{Comparison of training results with SGD, PHS and Goal-Oriented approaches for the FashionMNIST dataset.} \label{table:accuracies_fashionmnist}
\end{table}

\begin{figure*}
\centering
    \begin{subfigure}{0.4\linewidth}
    \centering
    \includegraphics[width=\linewidth]{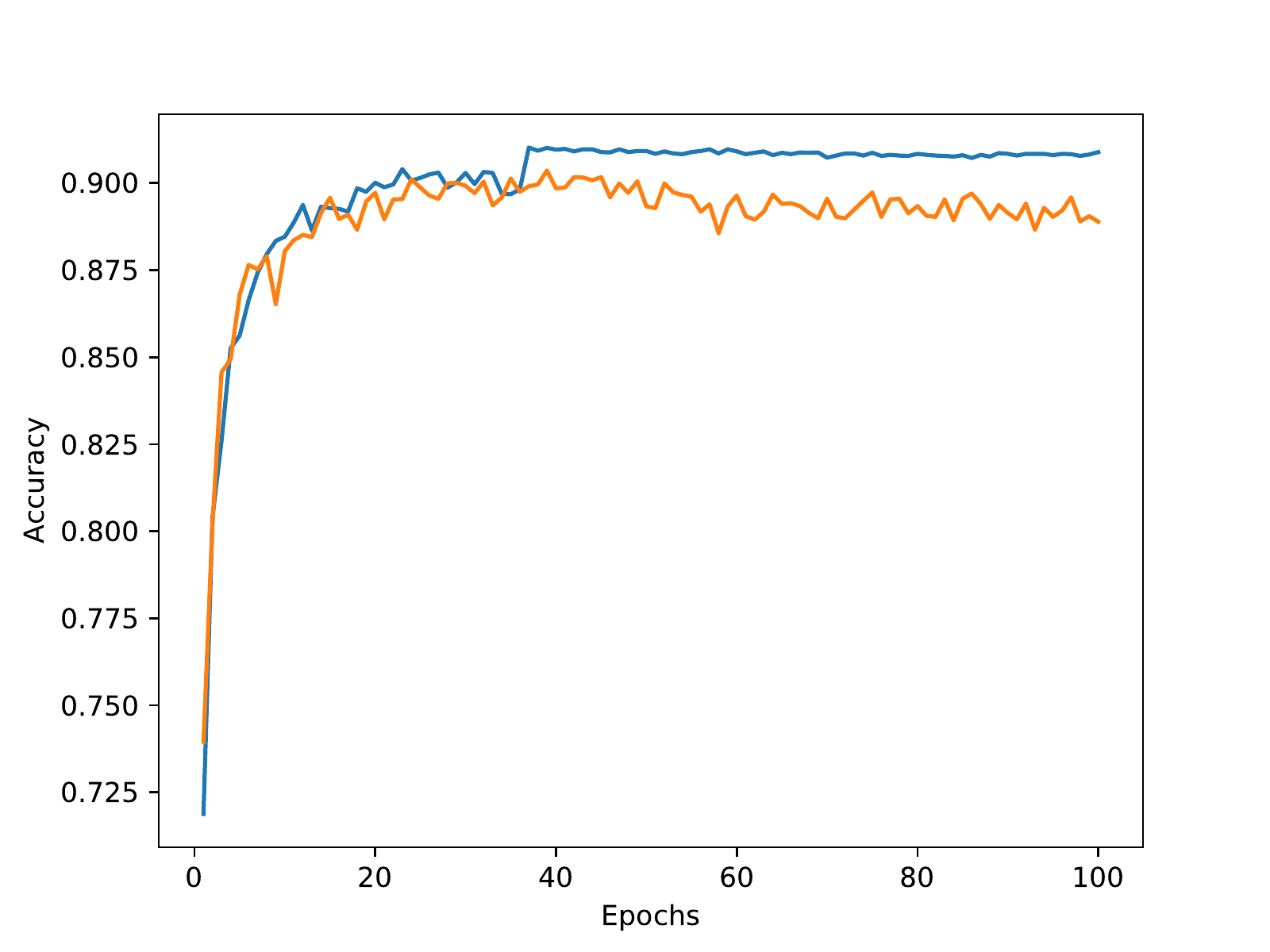}
    \caption{$\alpha=0.1$, friction = 0.1, mass = 10 on Fashion-MNIST.}
    \label{fig:acc_fmnist_fric01_lr01_mass10}
\end{subfigure}
\hspace{0.2cm}
\begin{subfigure}{0.4\linewidth}
    \centering
    \includegraphics[width=\linewidth]{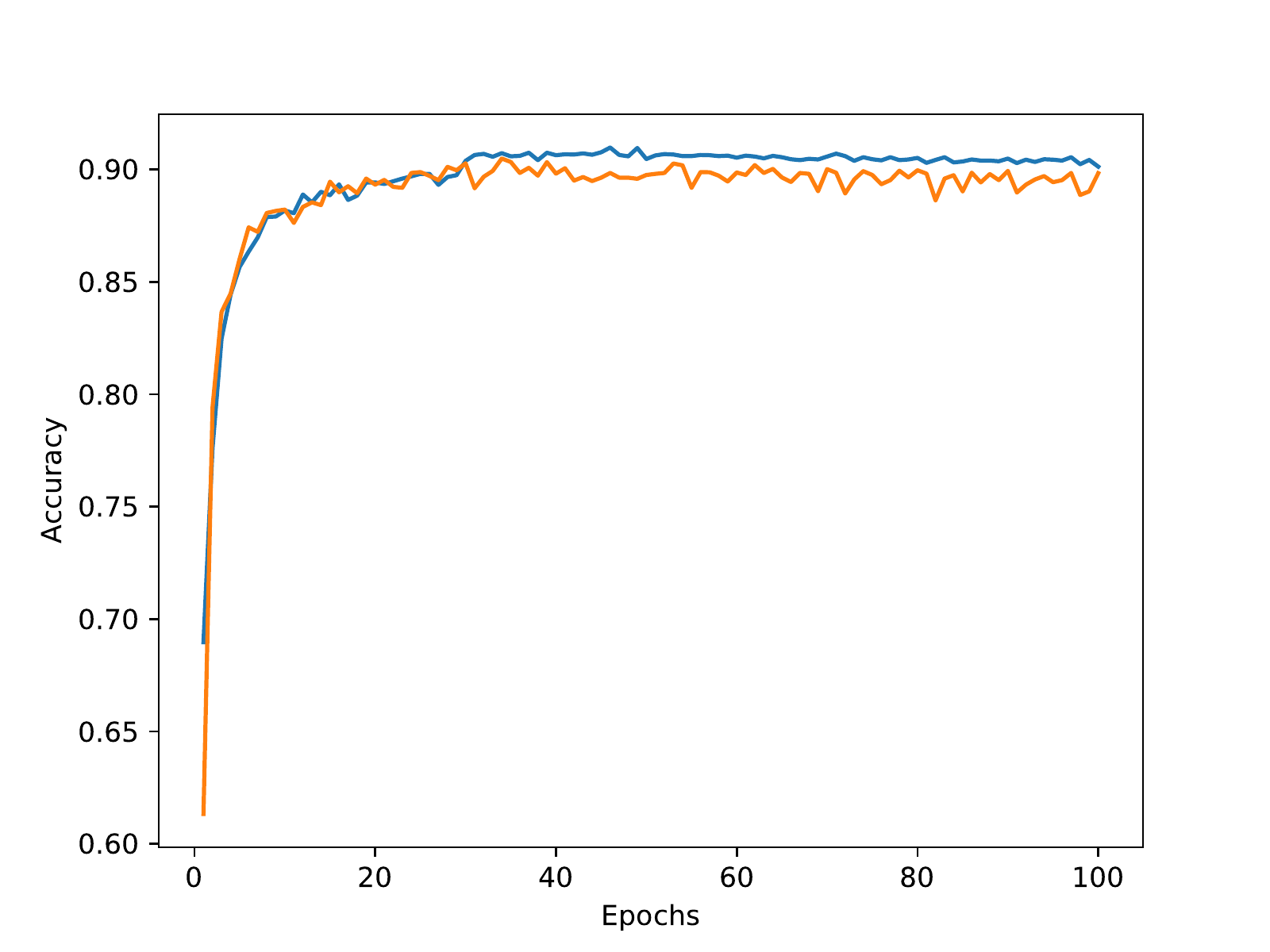}
    \caption{$\alpha=0.01$, friction = 1, mass = 0.1 on Fashion-MNIST. }
    \label{fig:acc_fmnist_fric1_lr001_mass01}
\end{subfigure}
    \begin{subfigure}{0.4\linewidth}
    \centering
    \includegraphics[width=\linewidth]{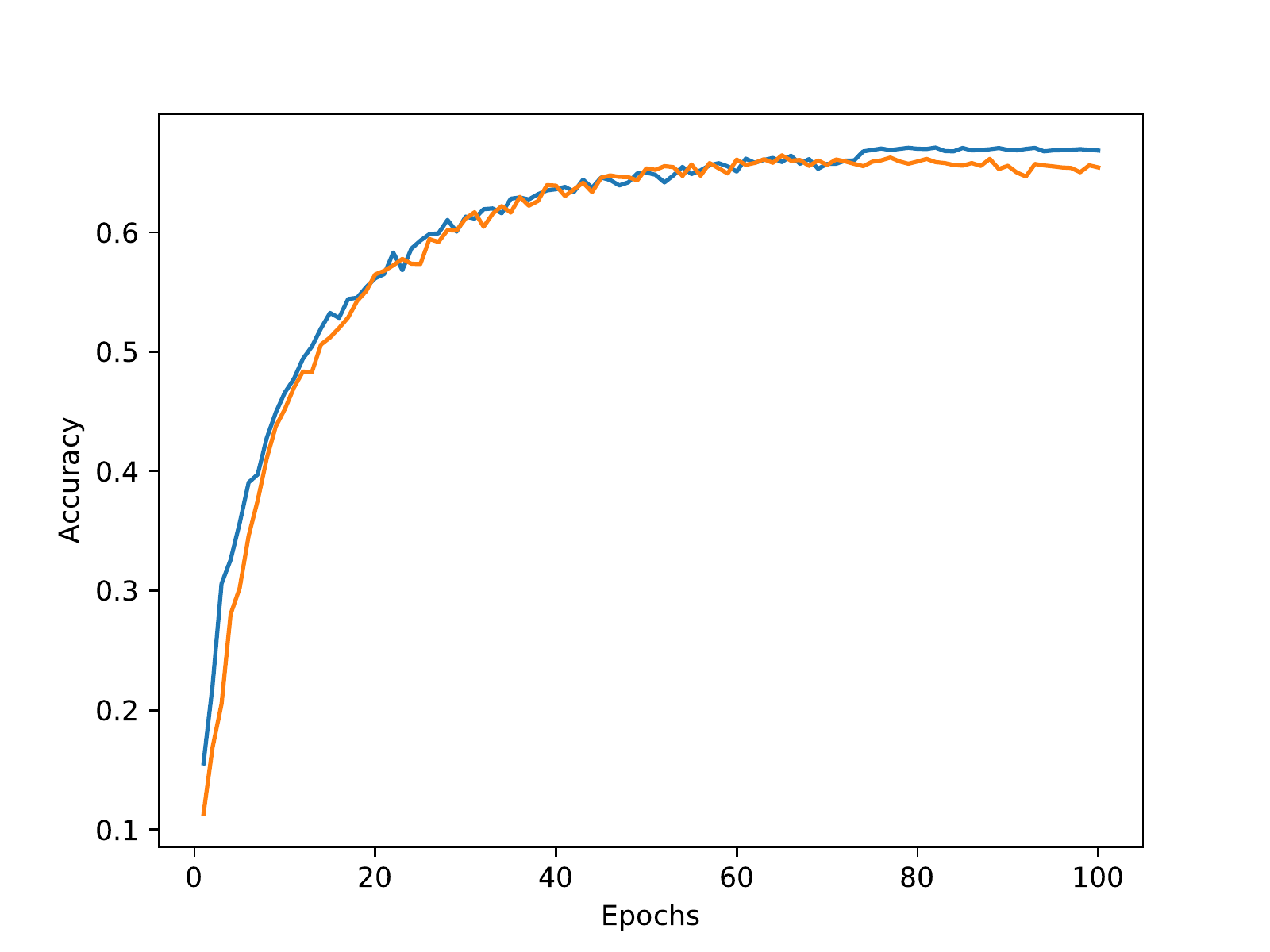}
    \caption{$\alpha=0.1$, friction = 0.1, mass = 100 on CIFAR-10.}
    \label{fig:acc_cifar10_fric01_lr01_mass100}
\end{subfigure}
\hspace{0.2cm}
\begin{subfigure}{0.4\linewidth}
    \centering
    \includegraphics[width=\linewidth]{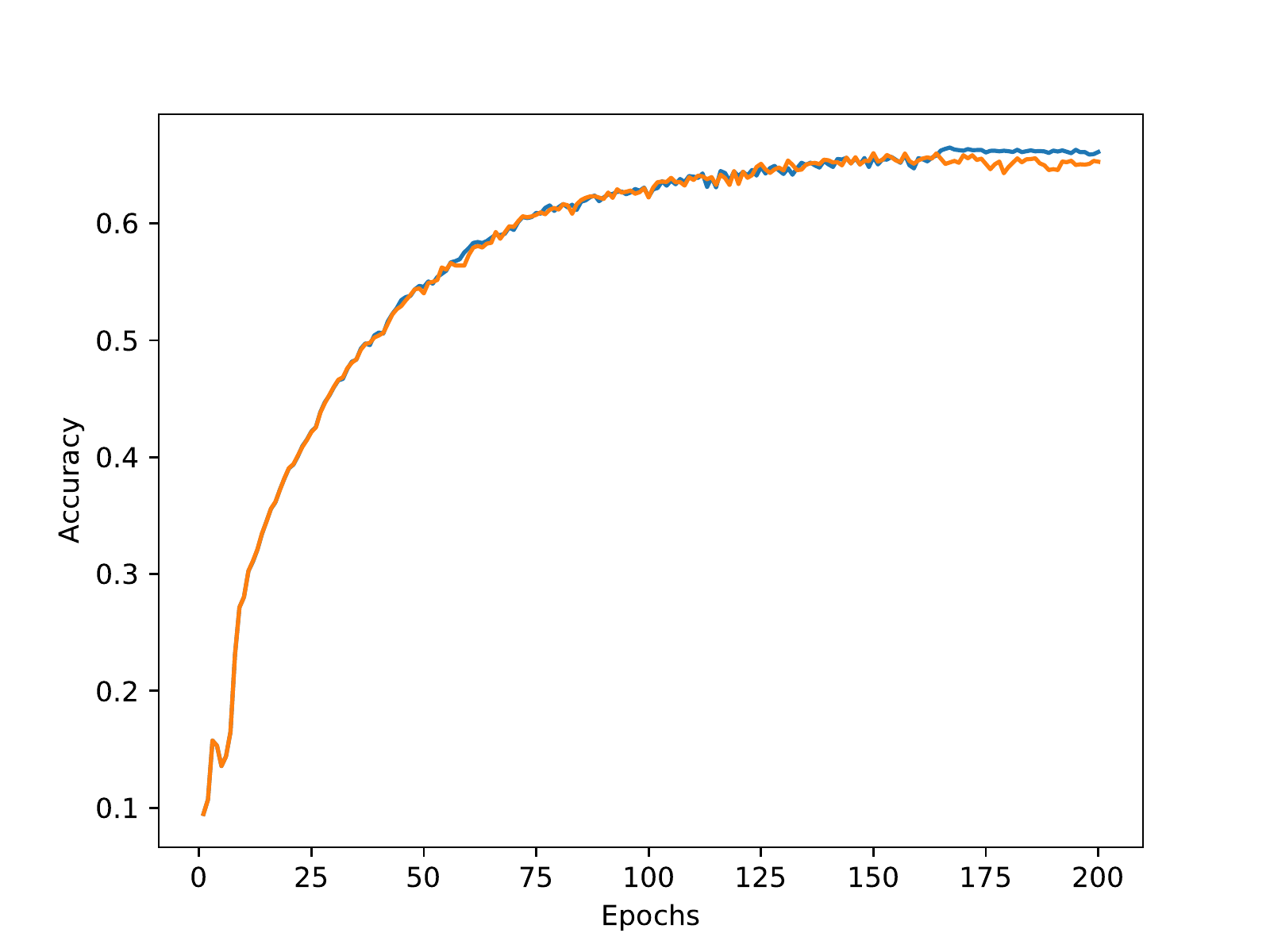}
    \caption{$\alpha=0.01$, friction = 0.1, mass = 25 on CIFAR-10.}
    \label{fig:acc_cifar10_fric01_lr001_mass25}
\end{subfigure}
\caption{History of the accuracies over the epochs depending on the choosable hyperparameters learning rate $\alpha$, friction and mass. PHS in orange, Goal-oriented approach in blue.}
\label{fig:hist_acc}
\end{figure*}

As can be seen in figure \ref{fig:acc_fmnist_fric01_lr01_mass10}, the accuracy of the method is consistantly improved by breaking after reaching the goal, and the subsequent occurrence of overfitting (as happens with the PHS) is avoided. The increase in test accuracy lies around and in many cases above 0.5\%  throughout parameter settings and the two data sets employed, as documented in Table \ref{table:accuracies_cifar10} for CIFAR10 and Table \ref{table:accuracies_fashionmnist} for FashionMNIST. 

The history of the test accuracy over the iteration count of the optimization procedure is shown in Figure \ref{fig:hist_acc} for two example configurations of each dataset. As we observe, the sudden 'breaking' exploits a local minimum better and avoids overfitting (as it can be especially seen in figure \ref{fig:acc_fmnist_fric01_lr01_mass10}), i.e. the decrease of the ordinary PHS method in the further pursuit of the optimization. Interestingly, this hints that overfitting rather is a 'global' phenomenon associated with ongoing exploration, whereas exploitation of the local minimum seems less beset from overfitting issues. This is consistent with our observation that the training loss after 'breaking' quickly converges, whereas the training loss for SGD or PHS is further reduced. This suggest that the onset of overfitting could thus  also be a useful triggering event for 'breaking' instead of goal orientation, as employed here.

\section{Discussion and Outlook}
\label{sec:discussion}

In our paper, we have introduced a new goal oriented strategy for the training of deep neural networks. By the physics-motivated interpretation of momentum in a port Hamiltonian framework, we explained how different settings for the friction / dissipation correspond to an exploration or exploitation phase in the progress of optimization. By switching from exploration to exploitation when  a certain minimal reduction of the loss function of a deep neural network is achieved, we obtain improved classification accuracy of image classification networks as compared with simple stochastic gradient descent or a momentum based optimization with fixed friction.

The outlined strategy can be extended in several ways. First, for the case where the minimal reduction is never achieved for a long time, the exploitation phase could be executed nevertheless starting from the best parameter setting found so far, or the target could be adjusted. This will robustify our algorithm. Second, after  a first exploitation phase, a re-acceleration could be executed, e.g. by an external force or 'port', so that multiple promising local minima can be visited.

\vspace{.3cm}

\noindent \textbf{Acknowledgements:} The authors thank Onur T.\ Doganay, Kathrin Klamroth, Matthias Rottmann and Claudia Totzeck for interesting discussions.  This work is partially funded by the German Federal Ministry for
Economic Affairs and Climate Action, within the project “KI Delta
Learning”, grant no. 19A19013Q. 
\printbibliography

\end{document}